\documentclass[9pt,shortpaper,twoside,web]{ieeecolor}
\usepackage{generic}
\usepackage{cite}
\usepackage{amsmath,amssymb,amsfonts}
\usepackage{algorithmic}
\usepackage{algorithm}
\usepackage{amsmath}
\usepackage{graphicx}
\usepackage{textcomp}
\usepackage{subcaption}
\usepackage{multirow}
\def\BibTeX{{\rm B\kern-.05em{\sc i\kern-.025em b}\kern-.08em
    T\kern-.1667em\lower.7ex\hbox{E}\kern-.125emX}}
\begin{document}
\title{ECGMamba: Towards Efficient ECG Classification with BiSSM}
\author{Yupeng Qiang, Xunde Dong, Xiuling Liu,  \IEEEmembership{Member, IEEE}, Yang Yang, Yihai Fang, and Jianhong Dou
\thanks{This work was supported by the National Natural Science Foundation of China (Nos. 62003141 and U20A20224); the Natural Science Foundation of Guangdong Province, China (No. 2021A1515011598); the Fundamental Research Funds for the Central Universities (No. 2022ZYGXZR023).(Corresponding author: Xunde Dong.)}
\thanks{Yupeng Qiang is with the School of Automation Science and Engineering, South China University of Technology, Guangzhou 510641 China(e-mail: auqyp@mail.scut.edu.cn).}
\thanks{Xunde Dong is with the School of Automation Science and Engineering, South China University of Technology, Guangzhou 510641 China (e-mail: audxd@scut.edu.cn).}
\thanks{Xiuling Liu is with the Key Laboratory of Digital Medical Engineering of Hebei Province, College of Electronic and Information Engineering, Hebei University, Baoding, 071002, China (e-mail: liuxiuling121@hotmail.com).}
\thanks{Yang Yang is with the School of Automation Science and Engineering, South China University of Technology, Guangzhou 510641 China (e-mail: 202221018052@scut.edu.cn).}
\thanks{Yihai Fang is with Zhujiang Hospital of Southern Medical University, Guangzhou, 510280, China (e-mail: fang-yihai@163.com).}
\thanks{Jianhong Dou is with General Hospital of Southern Command, Guangzhou, 510030, China (e-mail: doujianhong@hotmail.com).}
}

\maketitle

\begin{abstract} 
Electrocardiogram (ECG) signal analysis represents a pivotal technique in the diagnosis of cardiovascular diseases. Although transformer-based models have made significant progress in ECG classification, they exhibit inefficiencies in the inference phase. The issue is primarily attributable to the secondary computational complexity of Transformer's self-attention mechanism, particularly when processing lengthy sequences. To address this issue, we propose a novel model, ECGMamba, which employs a bidirectional state-space model (BiSSM) to enhance classification efficiency. ECGMamba is based on the innovative Mamba-based block, which incorporates a range of time series modeling techniques to enhance performance while maintaining the efficiency of inference. The experimental results on two publicly available ECG datasets demonstrate that ECGMamba effectively balances the effectiveness and efficiency of classification, achieving competitive performance. This study not only contributes to the body of knowledge in the field of ECG classification but also provides a new research path for efficient and accurate ECG signal analysis. This is of guiding significance for the development of diagnostic models for cardiovascular diseases.
\end{abstract}

\begin{IEEEkeywords}
\textit{electrocardiogram (ECG), ECG classification, state-space model, Mamba.}
\end{IEEEkeywords}

\section{Introduction}
\label{sec:introduction}
The field of multilead electrocardiogram (ECG) diagnostics has witnessed a rapid growth in the number of complex models as deep learning techniques have become popular. Initially, researchers relied on convolutional neural networks (CNNs) \cite{strodthoff2019detecting,liu2019mfb,han2023automated,qiang2024automatic,jyotishi2023attentive,yao2020multi} and recurrent neural networks (RNNs) \cite{liu2019mfb,feng2019myocardial,prabhakararao2020myocardial,mahendran2021deep} for extracting discriminative features in ECG signals. Despite the success of these methods in feature extraction, they encounter limitations when dealing with dependencies in long sequences, particularly the limitations of RNNs in parallel processing. The advent of self-attentive mechanisms has ushered in a new era of breakthroughs in this field, enabling the direct computation of relationships between any two points in the input sequence, irrespective of their positional distance. The Transformer architecture has fully exploited this capability, emerging as a leading choice for modeling this specific time series of ECG data \cite{cheng2023msw, natarajan2020wide,zhang2023token}. Although the self-attention-based Transformer is capable of improving the performance of ECG classification, its inherent quadratic computational complexity results in inefficiency in the inference phase, particularly when dealing with long ECG sequences or large-scale datasets.

In order to address the balance between classification performance and inference efficiency, this study employs State Space Models (SSMs) \cite{gu2021efficiently} as the core of ECG classification. SSMs, such as S4 \cite{gu2021efficiently} and S5 \cite{smith2022simplified}, have been widely used as replacements for RNNs, CNNs, and Transformers in natural language processing tasks. SSMs benefit from its recursive nature, which enables natural inference efficiency and the handling of long-term dependencies efficiently through structured state matrices. In recent times, Mamba \cite{gu2023mamba} has emerged as a variant of SSM, which addresses the data invariance and time invariance issues of existing SSMs. This is achieved through the introduction of a data-dependent selection mechanism and an efficient hardware-aware design. This enables the model to extract key information and filter out irrelevant noise based on the input data, thus improving the performance of modeling temporal information.

In this study, we propose ECGMamba, which is the first to investigate the potential of BiSSM for the efficient classification of ECGs. The Mamba-based block\cite{gu2023mamba} is employed as a foundation, with the incorporation of other techniques, including residual connectivity \cite{he2016deep}, layer normalization \cite{ba2016layer}, and feed forward networks, to enhance the nonlinear representativeness and training stability of the model. 
The contributions of this paper are summarized as follows:

\begin{itemize}
	\item This study represents the first investigation into the potential of BiSSMs in ECG classification, effectively addressing the trade-off between classification performance and inference efficiency.
	
	\item We propose ECGMamba, which combines the basic Mamba-based block with other techniques, such as the layer normalization and feed-forward network, to further improve the temporal modeling capability while maintaining efficient inference.
	
	\item  The experimental results on two public datasets demonstrate that our proposed ECGMamba exhibits notable advantages in both effectiveness and efficiency.
\end{itemize}

The rest of the paper is organized as follows: In Section \ref{section2}, we review related research work. Section \ref{section3} elucidates the proposed methodology in detail. In Section \ref{section4}, we present the experimental procedures and the corresponding results. In Section \ref{section5}, an elaborate discussion is provided. Finally, we conclude our findings in Section \ref{section6}.

\section{Related Work}
\label{section2}
\subsection{Deep models for ECG classification}
To date, a number of approaches have emerged in the field of ECG signal classification, driven by deep learning techniques, with the aim of accurately identifying cardiac diseases. For instance, Strodthoff et al. proposed an algorithm for the detection and interpretation of myocardial infarction (MI) based on a fully convolutional neural network \cite{strodthoff2019detecting}. To further extract ECG features, Liu et al. \cite{liu2019mfb} presented a multiple-feature-branch convolutional bidirectional recurrent neural network. The extensive utilisation of residual networks (ResNet) in the domain of ECG has prompted Hannun et al. to develop a 34-layer ResNet model (CNN\_Hannun), which outperforms the average diagnostic proficiency of arrhythmia specialists. Based on the residual network, DenseNet introduces a denser connectivity mechanism. Qiang et al. \cite{qiang2024automatic} proposed a multi-channel dense attention network for myocardial infarction localisation and detection, which achieved better performance. Furthermore, with the growing prevalence of attention mechanisms in the field of deep learning, Yao et al. proposed an attention-based time-incremental convolutional neural network for multi-class arrhythmia detection of ECG signals \cite{yao2020multi}. Zhu et al. proposed a Squeeze-and-Excitation Residual Network (SE-ResNet) integrated model, which consists of two residual neural network modules with SE-Net, designed to classify cardiac abnormalities from 12-lead ECG signals. Furthermore, the Transformer model, which employs a multi-head attention mechanism, has been applied to the task of classifying ECG signals, demonstrating its excellent sequence modelling capability. Natarajan et al. developed a wide and deep Transformer model based on the 12-lead ECG, which combines manually-designed ECG features and automatically-learnt features from Transformer to features for a variety of cardiac abnormality classification tasks \cite{natarajan2020wide}.

\subsection{State Space Models}
The State Space Models (SSMs) is used to describe the dynamic change process consisting of observed values and unknown internal state variables. Gu et al. proposes a Structured State Space Sequence (S4) model, an alternative to the Transformer architecture that models long-range dependencies without using attention. The property of linear complexity of state space sequence lengths has received considerable research attention. Smith et al. improves S4 by introducing MIMO SSM and efficient parallel scanning into the S4 layer to achieve parallel initialization and state reset of the hidden layer. He et al. proposes introducing dense connection layers into SSM to improve the feature representation ability of shallow hidden layer states. Mehta et al. improves the memory ability ofthe hidden layer by introducing gated units on S4. Recently, Gu et al. proposes the general language model Mamba, which has better sequence modeling capabilities than Transformers and is linearly complex. In this work, a bi-directional SSM (BiSSM) is introduced on top of Mamba to improve the representation of contextual information in the hidden layer.

\section{Method}
\label{section3}
\begin{figure*}[htb!]
	\centering
	{\includegraphics[width=1.95\columnwidth]{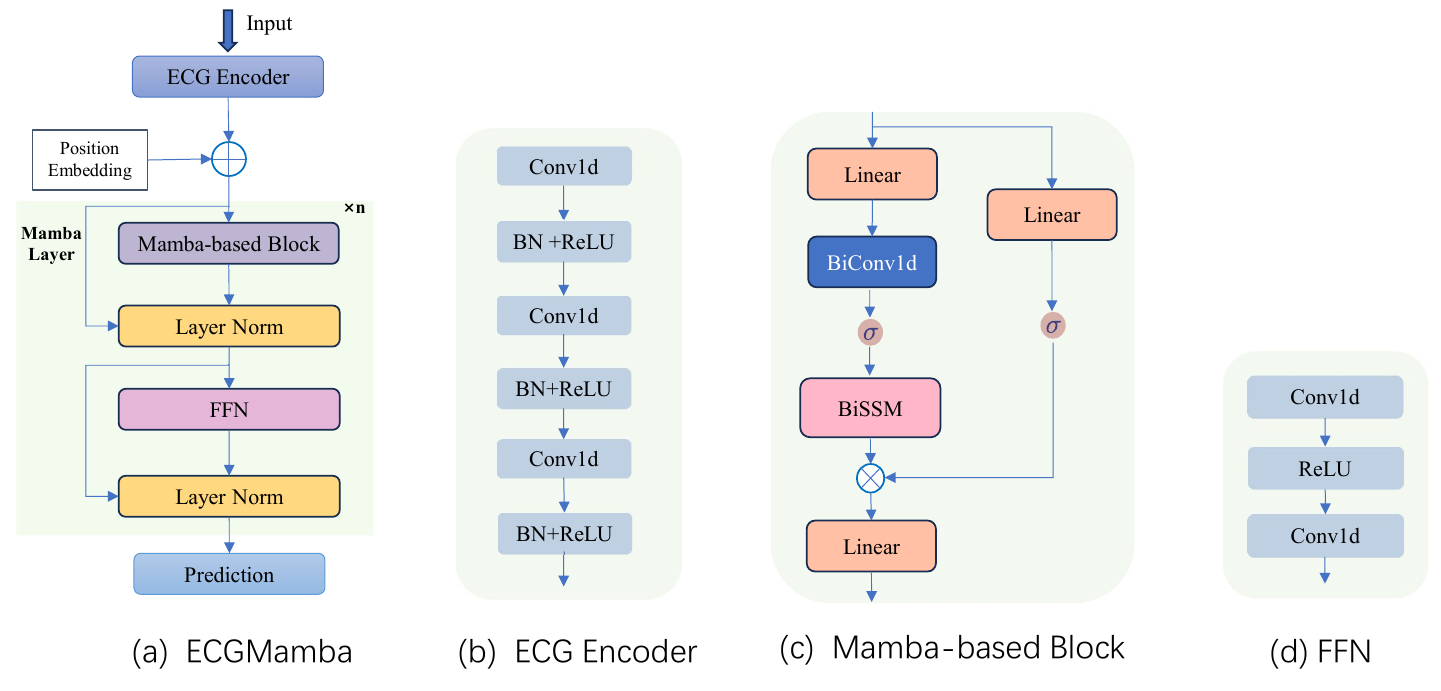}}
	\caption{Overall structure and detailed components of ECGMamba. FFN: Feed-Forward Network; Layer Norm: Layer Normalization; Conv1d: 1D convolution; BN: Batch Normalization; ReLU: Rectified Linear Unit; BiSSM: bidirectional SSM.}
	\label{ECGMamba}
\end{figure*}

This section will present the proposed ECGMamba framework, which is shown in Fig.~\ref{ECGMamba}. The framework comprises the ECG encoder and the Mamba layer. The Mamba layer primarily comprises the Mamba-based block (whose core is BiSSM) and a feed-forward network (FFN), which enables the model to effectively capture disease-specific information and continuous semantics from ECG signals.

\subsection{Basic Knowledge}
The SSMs is an efficient sequence modeling model that can capture the dynamic changes of data over time. Owing to the efficient sequence modeling capabilities, SSM has received widespread attention in various fields, e.g., video understanding and natural language process. A typical SSM consists of a state equation and an observation equation, where the state equation describes the dynamic changes within the system, and the observation equation describes the connection between the system state and observations. Given an input $x(t) \in R^D $  and a hidden state  $h(t) \in R ^N$, $ y(t) \in R^N $ is obtained mathematically through a linear ordinary differential equations (ODE) as follows:
\begin{flalign}
	\begin{split}
		h^{'}\left( t \right) &= Ah\left( t \right) +Bx\left( t \right) \\
		y\left( t \right) &=Ch\left( t \right)
	\end{split}
\end{flalign}
where $A \in R^{N \times N}$ and $B, C \in R^{N \times D}$ are learnable matrices, and N is the hidden state size. SSM is a continuous time series model, which is difficult to efficiently integrate into deep learning algorithms. 

Inspired by SSM, Mamba block discretizes ODEs to achieve computational efficiency. Mamba block discretizes the evolution parameter A and the projection parameter B by introducing a timescale parameter to obtain A and B. The formula is defined as follows:
\begin{flalign}
	\begin{split}
		\bar{A}&=\exp \left( \bigtriangleup A \right) \\
		\bar{B}&=\left( \bigtriangleup A \right) ^{-1}\left( \exp \left( \bigtriangleup B \right) -I \right) 
	\end{split}
\end{flalign}

After obtaining the discretized A and B, we rewrite Eq. 2 as follows:
\begin{flalign}
	\begin{split}
		h_t&=\bar{A}h_{t-1}+\bar{B}x_t\\
		y_t&=Ch_t
	\end{split}
\end{flalign}
and then the output is computed via global convolutiona as follows:
\begin{flalign}
	\begin{split}
		\bar{K}&=(C\bar{B},C\bar{A}\bar{B},\dots, C\bar{A}^L\bar{B})\\
		y &=x * \bar{K}
	\end{split}
\end{flalign}
where $L$ is the length of the input sequence $x$, and $\bar{K} \in R^L$ is a structured convolutional kernel.


\subsection{ECG Encoder}
Since ECG signal is of high resolution in the temporal domain, it is vital to efficiently extract temporal features per lead. Our default ECG encoder, as shown in Fig. 1b, consists of some temporal convolution blocks to encode each ECG signal into embeddings. Specifically, each temporal convolution block is composed of several 1-D convolution  (Conv1d) layers, batch normalization (BN) layers, and Rectified Linear Unit (ReLU) activation layers. 
The Conv1d layer is employed to extract temporal features from ECG signals and to capture local patterns. The BN layer improves the training speed and stability of the model by normalizing the input data. The ReLU layer introduces non-linear features, enhancing the expressive power of the model. The convolution operation is performed using a non-overlapping shift.

Furthermore, to facilitate the model to capture the dependencies between long-distance positions in the sequence, we introduce sine and cosine position encoding embedding as follows:
\begin{flalign}
	\begin{split}
		PE_{(pos,2i)}=sin(\frac{pos}{10000^{2i/d}} )\\
		PE_{(pos,2i+1)}=cos(\frac{pos}{10000^{2i/d}} )
	\end{split}
\end{flalign}
where pos represents the position in the sequence. i represents the dimension index of position encoding, i = 0, 1, ..., D-1. D represents the embedded dimension. Subsequently, we input the features  that encodes position information at each time step into the Mamba layer.

\subsection{Mamba Layer}
In this paper, we utilize the Mamba-based block for sequence classification tasks and introduce the Mamba layer, which primarily combines the Mamba-based block, as shown in Fig. 1c and FFN, as shown in Fig. 1d. In particular, this paper employs Mamba layers as the fundamental structural elements of the proposed architecture, rather than merely aggregating Mamba-based blocks in a sequential manner. The rationale behind this design is to integrate local features and global dependencies more effectively, thereby enhancing the accuracy and stability of sequence classification.

\subsubsection{Mamba-based Block}
The Mamba-based block structure is depicted in Fig. 1c. 
In order to aggregate forward and backward contextual semantic information, we construct a bidirectional SSM (BiSSM) component based on Mamba's unidirectional SSM component, inspired by LSTM(or unidirectional RNN) to BiLSTM(or BiRNN). The formal definition of the entire operation process is presented in Algorithm 1.


The Mamba-based block operates on an input $x_t \in R^{B\times L \times D}$ with dimensions of batch size $B$, sequence length $L$, and hidden dimension $D$ at each time step $t$. It begins by performing linear projections on the input $x_t$ with an expanded hidden dimension of $D$ to obtain $x$ and $z$ (Algo.1,2). These projections are then processed through a 1D convolution and a Swish \cite{} activation (Algo.4). The core of the block involves a BiSSM with parameters discretized based on the input (Algo.5-10). This discretized BiSSM, along with $x^{'}$, generates the state representation $\{y_i | i \in \{forward,backward\}\}$ (Algo.11). Finally, $y_i$ is combined with a residual connection from $z$ after applying Swish, and a final linear projection delivers the final output $y_t$ at time step $t$ (Algo.13,14). 
Overall, the Mamba-based block leverages input-dependent adaptations and BiSSM to process sequential information effectively. 

\begin{algorithm}[htpb]
	\caption{Mamba-based block}
	\label{BiSSM}
	\renewcommand{\algorithmicrequire}{\textbf{Input:}}
	\renewcommand{\algorithmicensure}{\textbf{Output:}}
	\begin{algorithmic}[1]
		\REQUIRE $x_t$: (B, L, D). 
		\ENSURE $y_t$ : (B, L, D).    
		\STATE x (B,M,E) $\leftarrow$ Linear($x_t$)
		\STATE z (B,M,E) $\leftarrow$ Linear($x_t$)
		\FOR {$i \in \{forward, backward\}$}
		\STATE $x^{'}$: (B, L, ED) $\leftarrow$ Swish(Conv1d$_{i}$(x))
		\STATE A$_i$: (D, N) $\leftarrow$ Parameter$_i$
		\STATE B$_i$: (B, L, N) $\leftarrow$  Linear($x^{'}$) 
		\STATE C$_i$: (B, L, N) $\leftarrow$  Linear($x^{'}$) 
		\STATE $\triangle_{i}$: (B, L, D)  $\leftarrow$ Softplus(Parameter$_i$+Broadcast(Linear(x$^{'}$)))
		\STATE $\bar{A}_{i}$: (B, L, D, N) $\leftarrow$ discrete($\triangle_{i}$, A$_i$)
		\STATE $\bar{B}_{i}$: (B, L, D, N) $\leftarrow$ discrete($\triangle_{i}$,B$_i$)
		\STATE $y_i$: (B, L, ED) $\leftarrow$ SSM($\bar{A}_{i}$,$\bar{B}_{i}$,C$_i$)($x^{'}$) 
		\ENDFOR
		\STATE $y^{'}$: (B, L, ED) $\leftarrow$ $\sum_{i\in {\{forward, backward\}}}^{} y_i $ $\otimes$ Swish(z)
		\STATE $y_t$: (B, L, D) $\leftarrow$ Linear($y^{'}$)
		\RETURN $y_t$
	\end{algorithmic}
\begin{flushleft}
	Parameter Description: \\ $N$: an SSM state expansion factor;  \\ $K$: a kernel size for the convolution;\\
	 $E$: a block expansion factor for input and output linear projections. \\
	 Softplus: an activation function.\\
\end{flushleft}
\end{algorithm}



\subsubsection{FFN}
Different from the 2-layer dense network in Transformer, we use a 2-layer Conv1d with ReLU activation as the FFN. This is because in the ECG classification task, neighbouring hidden states are more closely related in capturing the temporal dynamics of the ECG signal. 


Finally, the Layer Normalization is used after both Mamba block and FFN to stabilize the training process and reduce the risk of over-fitting.

\subsection{Memory-Efficiency} 
To avoid out-of-memory problems and achieve lower memory usage when dealing with long
sequences, Mamba-based block chooses the same recomputation method as Mamba. For the intermediate states of size $(B,L,D,N)$ to calculate the gradient, Mamba-based block recomputes them at the network backward pass. For intermediate activations such as the output of activation functions and convolution, Mamba-based block also recomputes them to optimize the GPU memory requirement, as the activation values take a lot of memory but are fast for recomputation.

\subsection{Computation-Efficiency}
SSM in Mamba-based block and the self-attention mechanism in Transformer both plays an important role in modeling global contextual semantic information. However, the self-attention mechanism is quadratic (O($L^2$) to sequence length L, and SSM is linear to sequence length L (O($L$)). The computational efficiency makes Mamba-based block well-suited for  modeling long sequences.
\section{Experiments and Results}
\label{section4}
\subsection{Basic Setting}
All the code used in this paper is implemented using PyTorch framework. The experiments were conducted on an Inter I7-13700K server equipped with NVIDIA RTX 4090 24GB GPU. 
The optimizer used was AdamW with a warmup learning rate strategy, and the batch size is 64. For the parameters of the Mamba-based block, the SSM state expansion factor $N$ is 32, the kernel size $K$ for BiConv1d is 4, and the block expansion factor $E$ for linear projections is 2. 

\subsection{Datasets}

\subsubsection{PTB-XL Dataset}
The PTB-XL dataset \cite{wagner2020ptb} comprises a multi-label collection of 21,799 clinical 12-lead ECG recordings from 18,869 patients, each lasting 10 seconds. These recordings were classified into five distinct classes: Normal (NORM), Myocardial Infarction (MI), ST/T Change (STTC), Conduction Disturbance (CD), and Hypertrophy (HYP). The publisher partitioned the PTB-XL database into 10 subsets, organized according to inter-patient paradigms specifically tailored for training, validation, and testing purposes. In accordance with previous studies and to facilitate fair comparisons, we adopted to the dataset division protocol recommended by the PTB-XL dataset publisher \cite{wagner2020ptb}. Specifically, subsets 1-8 were utilized for training (17084), subset 9 for validation (2146), and subset 10 for testing  (2158) within our study. It is pertinent to emphasize that our analysis employed the 100Hz data version and did not involve any preprocessing steps on the PTB-XL database.

\subsubsection{CPSC2018 Dataset}
This multi-label dataset was derived from the China Physiological Signal Challenge 2018 \cite{liu2018open}, which contains 6,877 12-lead ECG recordings ranging from 6 to 60 seconds with a sampling frequency of 500 Hz. These recordings were classified into nine distinct classes: Normal (NORM), Atrial fibrillation (AFIB), First-degree atrioventricular block (I-AVB), Left bundle brunch block (LBBB), Right bundle brunch block (RBBB), Premature atrial contraction(PAC),  Premature ventricular contraction(PVC),  ST-segment elevated(STE), ST-segment depression (STD). To ensure uniformity in the input signal lengths, we used simple data processing techniques  First, we reduce the sampling frequency of the data to 100 Hz. Secondly, signals longer than 10 seconds were cropped, while signals shorter than 10 seconds were padded with zeros to match the desired duration. To conduct inter-patient experiments, The paper employed the stratified sampling method \cite{kohavi1995study} to divide the data into 10 subsets, each containing records from different patients. We specify subsets 1 to 8 as the training set (5497), use subset 9 as the validation set (690), and subset 10 as the test set (690).

\subsection{Evaluation Metrics}
In this study, we comprehensively evaluate the performance of the model using three metrics: Area Under the Curve($AUC$), $F1$-score ($F1$), and Accuracy ($Acc$). The $AUC$ metric measures the overall performance of the classification model. A higher value indicates a better ability of the model to discriminate between positive and negative samples. The $F1$ metric represents the balance between precision and recall. It is a harmonic mean for precision and recall and is commonly used for evaluating the classification performance of a model on a multi-category imbalanced data set. The $Acc$ metric reflects the ability of a classifier or model to determine the correctness of all samples, which is defined as the ratio of correctly classified samples to the total number of samples.

\subsection{Results}
To evaluate the performance of ECGMamba, this study conducted two experiments using the datasets mentioned above. The experiments are as follows:  1) Conduct 5-category classification experiments on the PTB-XL dataset, comprising NORM, MI, STTC, CD, and HYP categories. 2) Conduct 9-category classification experiments on the CPSC2018 dataset, including NORM, AFIB, IAVB, LBBB, RBBB, PAC, PVC, STE, and STD categories.

Tables~\ref{R1} and~\ref{R2} demonstrate a comprehensive comparison of the ECGMamba model with existing studies in terms of methods, performance metrics, number of parameters (Param.), and number of floating point operations (FLOPs). On PTB-XL and CPSC2018, two widely used benchmark datasets, the ECGMamba model demonstrates excellent performance.

On the PTB-XL dataset, the ECGMamba model achieves an AUC of 91.67\%, an F1 value of 75.28\%, and an accuracy of 86.08\%, whose overall performance outperform models in existing studies. Compared with ASTL-Net, an existing method that performed well in terms of composite metrics, ECGMamba improved by 0.36\%, 1.70\%, and 3.58\% in terms of AUC, F1, and Acc metrics, respectively. In addition, although the number of parameters in ECGMamba is twice that of ASTL-Net, the FLOPs of ECGMamba are only 1/9th of that of ASTL-Net, showing its efficiency in performing the task.

On the CPSC2018 dataset, the ECGMamba model has an AUC of 94.26\%, an F1 score of 78.09\%, and an accuracy of 95.39\%, and the overall metrics are still superior to existing methods. Compared to DMES-net, a model with excellent overall metrics in this comparative study, ECGMamba improves 15.99\% on Acc and 1.39\% on F1, showing a significant advantage in its ability to judge the correctness of all samples, despite being slightly lower on AUC by 1.44\%. It is worth noting that the number of parameters and the number of floating-point operations of ECGMamba are about 3/4 and 1/8 of those of DMES-net, respectively.

By analyzing Tables~\ref{R1} and~\ref{R2}, we can find that the number of parameters and floating-point operations of ECGMamba are similar to that of VIT, but ECGMamba is much higher than it in terms of performance metrics. AUC, F1 and Acc are improved by at least 10\%, 20\% and 20\% on each dataset. In addition, despite the fact that ECGMamba has 6 times more parameters than LSTM, it has a significant improvement in FLOPs and performance metrics compared to it, where the FLOPs are only 1/8 of LSTM. On the PTB-XL dataset, AUC, F1, and Acc are improved by 1.18\%, 4.63\%, and 5.93\%, respectively; on the CPSC2018 dataset, the AUC, F1 and Acc were improved by 7.56\%, 22.19\% and 12.19\%, respectively. Finally, compared to Transformer$_{Wd}$, the overall metrics of ECGMamba are improved, with the number of parameters and the number of floating-point operations being only 1/2 of its size, while AUC, F1, and Acc are improved by at least 2\%, 4\%, and 2\% on each dataset.

In summary, the ECGMamba model, a robust and efficient tool for ECG signal analysis, shows significant generalisation ability and performance consistency across multiple datasets. This achievement is mainly attributed to the design of the Mamba Block, which is capable of selectively remembering or forgetting information through the selective properties of the selective SSM, which is crucial for capturing complex dependencies in sequence data. In addition, the design of the selective SSM enables it to be as efficient as traditional discretised parameter-based RNNs in the inference process, thus achieving a good balance between resource efficiency and model performance.
\begin{table}[]
	\centering
	\caption{Contrastive experimental results on the PTB-XL database.}
	\label{R1}
	\begin{tabular}{cccccc}
		\hline
		 Model          & {AUC(\%)} &F1(\%) &Acc(\%) & Param. & FLOPs   \\ \hline
		LSTM           & 90.49  & 70.65  & 80.15 & \textbf{0.90M}  & 807.17M \\
		Resnet50                  & 91.01  & 70.43  & 80.75 & 16.03M & 821.72M \\
		DMES-net       & 90.72  & 70.72  & 80.51 & 7.08M  & 791.90M \\
		CRT & {89.22} & {68.43} & {87.81} & {18.00M} & {401.80M} \\
		ATI-CNN        & 91.05  & 71.75  & 82.32 & 5.00M  & 574.36M \\
		VIT            & 83.16  & 49.50  & 55.29 & 4.14M  & 189.20M \\
		SPNV2                     & ——     & \textbf{76.20}  & 80.50 & 7.81M  & 199.14M \\
		ASTL-net       & 91.31  & 73.58  & 82.69 & 2.24M  & 910.23M \\
		SE-Resnet       & 89.33  & 69.77  & 87.46 & 16.77M  & 821.72M \\
		Transformer$_{Std}$ & 88.16  & 71.21  & 83.14 & 11.02M & 207.14M \\
		Transformer$_{Wd}$  & 89.71  & 72.54  & 84.29 & 12.33M & 269.43M \\
		ECGMamba      & \textbf{91.67}  & {75.28}  & \textbf{88.81} & 5.31M  & \textbf{101.33M} \\ \hline
	\end{tabular}
\end{table}

\begin{table}[]
	\centering
	\caption{Contrastive experimental results on the CPSC2018 database.}
	\label{R2}
	\begin{tabular}{cccccc}
		\hline
		Model                & {AUC(\%)} &F1(\%) &Acc(\%) & Param.         & FLOPs   \\ \hline 
		Resnet50            & 94.80  & 70.90  & 91.40 & 16.03M          & 821.72M \\
		LSTM                & 87.70  & 56.90  & 83.20 & \textbf{0.90M} & 807.17M \\
		VIT                 & 83.57  & 57.19  & 76.00 & 4.14M          & 189.20M \\
		DMES-net & \textbf{96.70} & {77.70} & {79.40} & {7.08M} & {791.90M} \\
		SPNV2     & ——       & 72.70  & 78.60 & 7.81M          & 199.14M \\
		SE-Resnet & 93.51  & 77.28  & 85.80 & 16.77M         & 821.72M \\
		Joint-net & 91.51  & 62.29  & 66.18 & ——             & ——      \\
		MVMS-net            & 95.33  & 77.22  & 74.64 & 2.24M          & 900.16M \\
		Transformer$_{Std}$     & 91.09  & 73.61  & 93.96 & 11.02M         & 207.14M \\
		Transformer$_{Wd}$      & 92.91  & 75.54  & 94.46 & 12.33M         & 269.43M \\
		ECGMFormer          & 95.26  & \textbf{79.09}  & \textbf{95.39} & 5.31M          & \textbf{101.33M} \\ \hline
	\end{tabular}
\end{table}

\subsection{t-SNE Visualisation}
In order to effectively visualize the performance of ECGMamba, we utilise the t-distributed stochastic neighbourhood embedding (t-SNE) \cite{cieslak2020t} technique to map the high-dimensional features extracted by ECGMamba from the ECG. As illustrated in Fig. 2, each data point in the graph represents a sample. It is evident that there is a notable clustering of distinct ECG patterns in the two classification tasks. Nevertheless, it is crucial to acknowledge that achieving complete separation of multiple distinct disease states in a patient-to-patient experiment remains a formidable challenge. Two major factors contribute to this challenge. Firstly, significant individual differences between patients significantly affect the discriminatory power of the model, which in turn affects the visualization results. Secondly, the limited availability of ECG data restricts the training of the model, which in turn affects the final level of performance.
\begin{figure*}[htpb]
	\centering
	\begin{subfigure}{0.49\linewidth}
		\centering
		\includegraphics[width=\columnwidth]{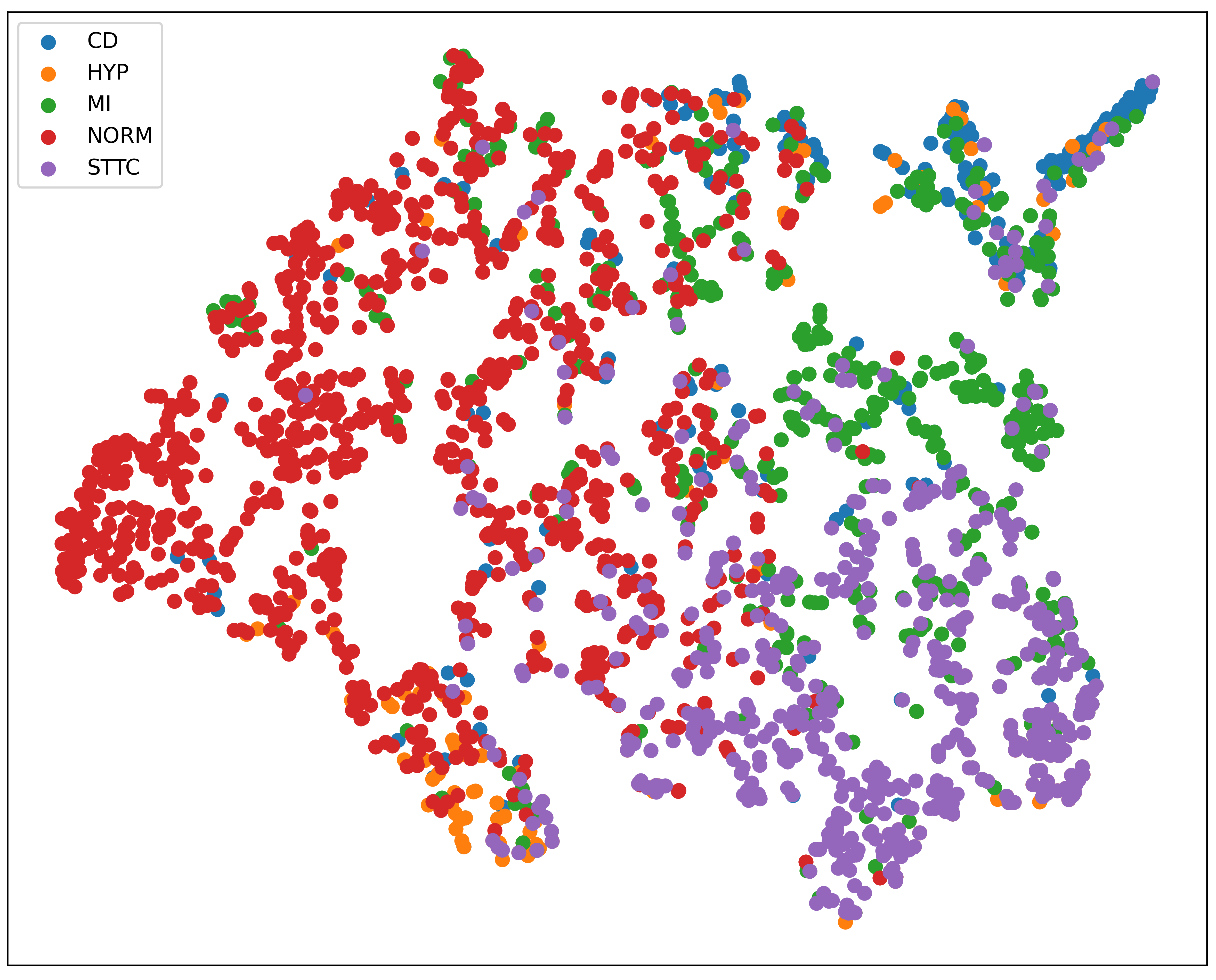}
		\caption{t-SNE visualization on PTB-XL task.}
		\label{}
	\end{subfigure}
	\begin{subfigure}{0.45\linewidth}
		\centering
		\includegraphics[width=\columnwidth]{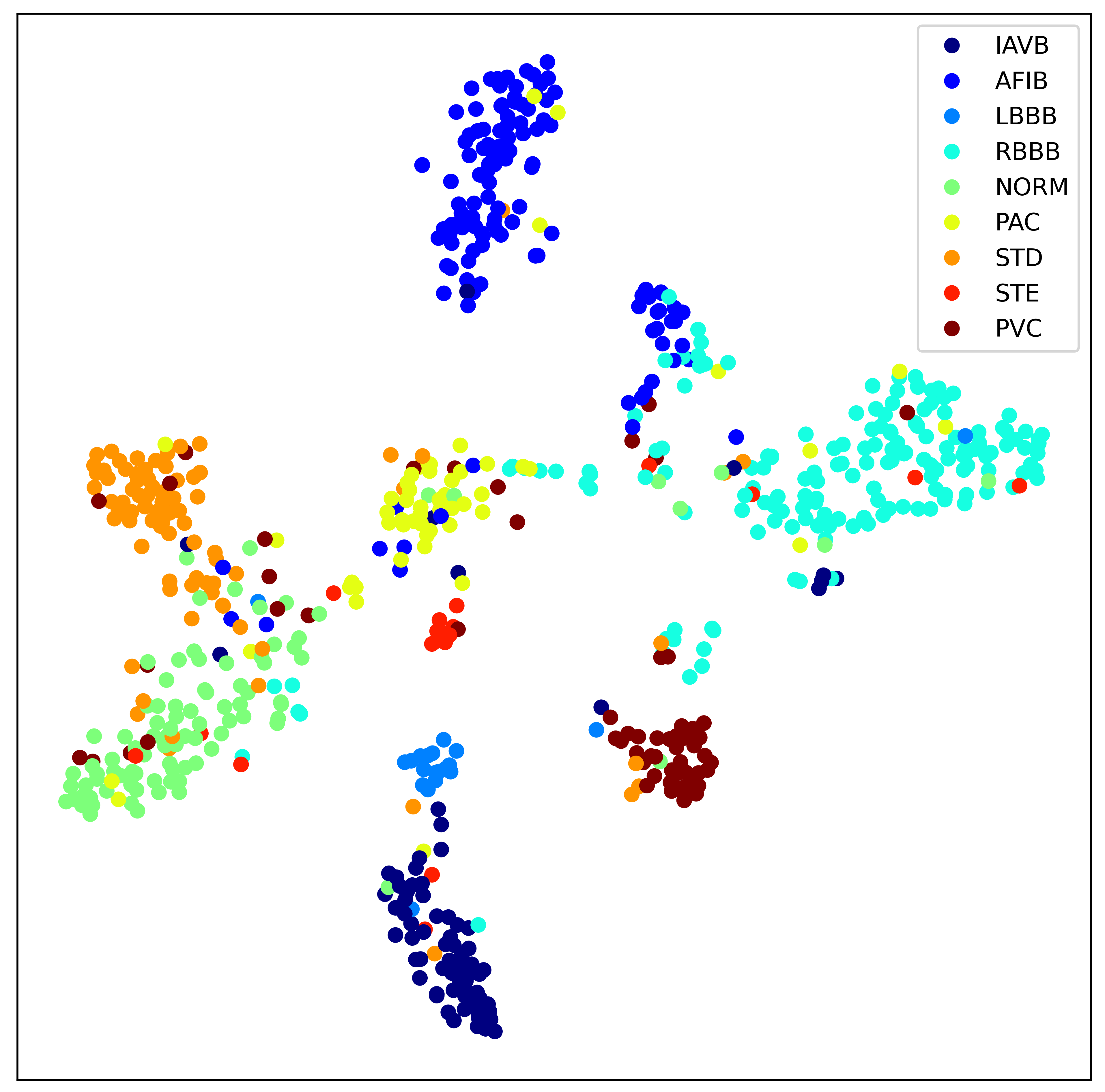}
		\caption{t-SNE visualization on CPSC2018 task.}
		\label{}
	\end{subfigure}
	\caption{t-SNE visualization of ECGMamba.}
	\label{tsne}
\end{figure*}
\section{Discussion}
\label{section5}

\subsection{The number of Mamba Layers}
In order to investigate the specific impact of the number of Mamba layers on the model performance in greater depth, this study compares the model performance under different Mamba layer configurations. The experimental results as shown in Fig.~\ref{D1} demonstrate that the model attains optimal performance when the number of Mamba layers is set to eight. Additionally, it was observed that the performance of the model does not continuously improve with the increase in the number of Mamba layers. This key finding indicates that in the process of model design and optimisation, it is crucial to seek an optimal balance between model complexity and performance. This is to ensure that the model can process various features efficiently while also maintaining efficient operation under limited computational resources. In particular, the number of parameters and computational complexity of the model increase in direct proportion to the number of Mamba layers. This may result in an increase in the time required for training and inference, but may also lead to a significant improvement in model performance.

\begin{table}[]
	\centering
	\caption{The results of the different Mamba Layers. The best results are bold.}
	\label{D1}
	\begin{tabular}{ccccccc}
		\hline
		 & \multicolumn{3}{c}{PTB-XL} & \multicolumn{3}{c}{CPSC2018} \\ \cline{2-7}
		N  & AUC    & F1     & Acc    & AUC    & F1     & Acc     \\ \hline
		1  & 0.8922 & 0.6877 & 0.8391 & 0.9115 & 0.6683 & 0.916  \\
		2  & 0.9006 & 0.6825 & 0.8803 & 0.9149 & 0.6984 & 0.9302 \\
		3  & 0.9017 & 0.6954 & 0.8853 & 0.9174 & 0.721  & 0.9373 \\
		4  & 0.9037 & 0.7167 & \textbf{0.8928} & 0.9099 & 0.7341 & 0.9397 \\
		5  & 0.8997 & 0.7069 & 0.8876 & 0.9258 & 0.7506 & 0.9427 \\
		6  & 0.9067 & 0.7157 & 0.8902 & 0.9087 & 0.7129 & 0.9358 \\
		7  & 0.8992 & 0.7012 & 0.8873 & 0.9190  & 0.7263 & 0.9381 \\
		8 & \textbf{0.9167} & \textbf{0.7528} & {0.8881} & \textbf{0.9526} & \textbf{0.7909} & \textbf{0.9539} \\
		9  & 0.9030  & 0.7040  & 0.8874 & 0.9213 & 0.7427 & 0.9422 \\
		10 & 0.9042 & 0.7148 & 0.8899 & 0.9233 & 0.7242 & 0.9374 \\ \hline
	\end{tabular}
\end{table}

\subsection{The contributions of each component}
In order to ascertain the contribution of each component in the architecture, an ablation study was conducted. The results are presented in Table~\ref{D2}. The following section presents each variant and analyses its respective impact.
\subsubsection{ECG Encoder}
The experimental results demonstrate that utilising the ECG encoder for the initial feature extraction of long time series as a means of providing initialisation information for the Mamba Layer can markedly enhance the overall performance of the model. This indicates that when processing time series data, CNN is capable of effectively capturing local features and patterns, thereby providing valuable information for the subsequent Mamba Layer.
\subsubsection{LN}
Following the Mamba blocks and FFN, the LN were applied. The experimental results demonstrate the efficacy of the LN in reducing the risk of overfitting and improving the performance of the model. LN helps to stabilise the training process by normalizing the layer outputs.
\subsubsection{FFN}
The experimental result indicated that the FFN played a role in enhancing the nonlinear modelling ability of the model and had a significant impact on the enhancement of model performance. The FFN was able to introduce more nonlinear transformations that helped the model to capture complex data relationships and thus improve performance.
\begin{table}[]
	\centering
	\caption{The results of the each component.}
	\label{D2}
	\begin{tabular}{ccccccc}
		\hline
		& \multicolumn{3}{c}{PTB-XL} & \multicolumn{3}{c}{CPSC2018} \\ \cline{2-7} 
		& AUC     & F1      & Acc    & AUC    & F1     & Acc    \\ \hline
		w/o ECG Encoder & 0.8719  & 0.7209  & 0.8512 & 0.9129 & 0.7604 & 0.9337 \\
		w/o LN  & 0.9141  & 0.7508  & 0.8843 & 0.9509 & 0.7887 & 0.9468 \\
		w/o FFN & 0.9028  & 0.7412  & 0.8794 & 0.9468 & 0.7762 & 0.9348 \\
		Proposed & \textbf{0.9167} & \textbf{0.7528} & \textbf{0.8881} & \textbf{0.9526} & \textbf{0.7909} & \textbf{0.9539} \\ \hline
	\end{tabular}
\end{table}

\section{Conclusion}
\label{section6}
In this study, we propose an innovative solution to the core challenge faced by the Transformer model in ECG signal analysis-namely, achieving efficient and accurate ECG classification: the ECGMamba model. The model effectively overcomes the efficiency bottleneck of the traditional Transformer model when dealing with long sequence data by incorporating the Mamba module with selective SSM.The introduction of ECGMamba not only significantly improves the performance of the ECG classification task, but also maintains the efficient inference speed. Experimental validation on two publicly available ECG datasets shows that ECGMamba exhibits excellent performance in the classification task, especially when processing long sequence data, with better accuracy and efficiency than existing methods. In addition, the competitiveness of ECGMamba in terms of performance metrics further demonstrates its potential for application in the field of ECG classification. The results of this study not only provide a new research direction in the field of ECG classification, but also contribute new ideas for the development of diagnostic models for cardiovascular diseases.

Although the ECGMamba model has demonstrated significant performance advantages, the study still has some limitations. Firstly, although the model has been successfully validated on three publicly available datasets, yielding commendable results and robustness, further testing on a wider range of datasets is necessary to determine its consistency and reliability in different environments and conditions. Second, the current discussion of the interpretability of ECGMamba is insufficient. Future research will focus on improving the interpretability of model predictions to increase clinician confidence and acceptance of automated ECG diagnostic systems.

\bibliographystyle{IEEEtran} 
\bibliography{ref}
\end{document}